\documentclass{article}

\usepackage[nonatbib,final]{nips_2016}

\usepackage[utf8]{inputenc} 
\usepackage[T1]{fontenc}    
\usepackage{hyperref}       
\usepackage{url}            
\usepackage{booktabs}       
\usepackage{amsfonts}       
\usepackage{nicefrac}       
\usepackage{microtype}      

\usepackage{graphicx}
\usepackage{caption}
\usepackage{subcaption}

\title{Review Networks for Caption Generation}

\author{
  Zhilin Yang,~~Ye Yuan,~~Yuexin Wu,~~Ruslan Salakhutdinov,~~William W. Cohen \\
  School of Computer Science \\
  Carnegie Mellon University \\
  \texttt{\{zhiliny,yey1,yuexinw,rsalakhu,wcohen\}@cs.cmu.edu}
}

\begin{document}

\maketitle

\begin{abstract}

We propose a novel extension of the encoder-decoder framework, called a \textit{review network}.
The review network is generic and can enhance any existing encoder-decoder model: in this paper, we consider RNN decoders with both CNN and RNN encoders. The review network performs a number of \textit{review} steps with attention mechanism on the encoder hidden states, and outputs a \textit{thought vector} after each review step; the thought vectors are used as the input of the attention mechanism in the decoder. We show that conventional encoder-decoders are a special case of our framework. Empirically, we show that our framework improves over state-of-the-art encoder-decoder systems on the tasks of image captioning and source code captioning.\footnote{Code and data available at \url{https://github.com/kimiyoung/review_net}.}

\end{abstract}

\section{Introduction}

\textit{Encoder-decoder} is a framework for learning a transformation from one representation to another. In this framework, an encoder network first encodes the input into a context vector, and then a decoder network decodes the context vector to generate the output. The encoder-decoder framework was recently introduced for sequence-to-sequence learning based on recurrent neural networks (RNNs) with applications to machine translation \cite{cho2014learning,sutskever2014sequence}, where the input is a text sequence in one language and the output is a text sequence in the other language. More generally, the encoder-decoder framework is not restricted to RNNs and text; e.g., encoders based on convolutional neural networks (CNNs) are used for image captioning \cite{vinyals2015show}.
Since it is often difficult to encode all the necessary information in a single context vector, an \textit{attentive encoder-decoder} introduces an attention mechanism to the encoder-decoder framework. An attention mechanism modifies the encoder-decoder bottleneck by conditioning the generative process in the decoder on the encoder hidden states, rather than on one single context vector only.
Improvements due to an attention mechanism have been shown on various tasks, including machine translation \cite{bahdanau2014neural}, image captioning \cite{xu2015show}, and text summarization \cite{rush2015neural}.

However, there remain two important issues to address for attentive encoder-decoder models. First, the attention mechanism proceeds in a sequential manner and thus lacks global modeling abilities. More specifically, at the generation step $t$, the decoded token is conditioned on the attention results at the current time step $\tilde{\mathbf{h}}_t$, but has no information about future attention results $\tilde{\mathbf{h}}_{t'}$ with $t' > t$. For example, when there are multiple objects in the image, the caption tokens generated at the beginning focuses on the first one or two objects and is unaware of the other objects, which is potentially suboptimal. Second, previous works show that discriminative supervision (e.g., predicting word occurrences in the caption) is beneficial for generative models \cite{fang2015captions}, but it is not clear how to integrate discriminative supervision into the encoder-decoder framework in an end-to-end manner.

To address the above questions, we propose a novel architecture, \textit{the review network}, which extends existing (attentive) encoder-decoder models. The review network performs a given number of \textit{review} steps with attention on the encoder hidden states and outputs a \textit{thought vector} after each step, where the thought vectors are introduced to capture the global properties in a compact vector representation and are usable by the attention mechanism in the decoder.
The intuition behind the review network is to review all the information encoded by the encoder and produce vectors that are a more compact, abstractive, and global representation than the original encoder hidden states.

Another role for the thought vectors is as a focus for multitask learning. For instance, one can use the thought vectors as inputs for a secondary prediction task, such as predicting discriminative signals (e.g., the words that occur in an image caption), in addition to the objective as a generative model. In this paper we explore this multitask review network, and also explore variants with weight tying.



We show that conventional attentive encoder-decoders are a special case of the review networks, which indicates that our model is strictly more expressive than the attentive encoder-decoders. We experiment with two different tasks, image captioning and source code captioning, using CNNs and RNNs as the encoders respectively. Our results show that the review network can consistently improve the performance over attentive encoder-decoders on both datasets, and obtain state-of-the-art performance.

\section{Related Work}

The encoder-decoder framework in the context of sequence-to-sequence learning 
was recently introduced for learning transformation between text sequences~\cite{cho2014learning,sutskever2014sequence},
where RNNs were used for both encoding and decoding.
Encoder-decoders, in general, can refer to models that learn a representation transformation using two network components, an encoder and a decoder. Besides RNNs, convolutional encoders have been developed to address multi-modal tasks such as image captioning \cite{vinyals2015show}.
Attention mechanisms were later introduced to the encoder-decoder framework for machine translation, with attention providing an explanation of explicit token-level alignment between input and output sequences \cite{bahdanau2014neural}. In contrast to vanilla encoder-decoders, attentive encoder-decoders condition the decoder on the encoder's hidden states.
At each generation step, the decoder pays attention to a specific part of the encoder, and generates the next token based on both the current hidden state in the decoder and the attended hidden states in the encoder. 
Attention mechanisms have had considerable success in other applications as well, including image captioning \cite{xu2015show} and text summarization \cite{rush2015neural}.

Our work is also related to memory networks \cite{weston2014memory,sukhbaatar2015end}. Memory networks take a question embedding as input, and perform multiple computational steps with attention on the \textit{memory}, which is usually formed by the embeddings of a group of sentences. Dynamic memory networks extend memory networks to model sequential memories \cite{kumar2015ask}. Memory networks are mainly used in the context of question answering; the review network, on the other hand, is a generic architecture that can be integrated into existing encoder-decoder models. Moreover, the review network learns thought vectors using multiple review steps, while (embedded) facts are provided as input to the memory networks. Another difference is that the review network outputs a sequence of thought vectors, while memory networks only use the last hidden state to generate the answer. \cite{vinyals2015order} presented a processor unit that runs over the encoder multiple times, but their model mainly focuses on handling non-sequential data and their approach differs from ours in many ways (e.g., the encoder consists of small neural networks operating on each input element, and the process module is not directly connected to the encoder, etc). The model proposed in \cite{graves2016adaptive} performs a number of sub-steps inside a standard recurrent step, while our decoder generates the output with attention to the thought vectors.

\section{Model}

Given the input representation $\mathbf{x}$ and the output representation $\mathbf{y}$, the goal is to learn a function mapping from $\mathbf{x}$ to $\mathbf{y}$.  For example, image captioning aims to learn a mapping from an image $\mathbf{x}$ to a caption $\mathbf{y}$. For notation simplicity, we use $\mathbf{x}$ and $\mathbf{y}$ to denote both a tensor and a sequence of tensors. For example, $\mathbf{x}$ can be a 3d-tensor that represents an image with RGB channels in image captioning, or can be a sequence of 1d-tensors (i.e., vectors) $\mathbf{x} = (\mathbf{x}_1, \cdots, \mathbf{x}_{T_x})$ in machine translation, where $\mathbf{x}_t$ denotes the one-of-$K$ embedding of the $t$-th word in the input sequence of length $T_x$.

\begin{figure}
    \centering
    \begin{subfigure}[t]{0.28\textwidth}
        \includegraphics[width=\textwidth]{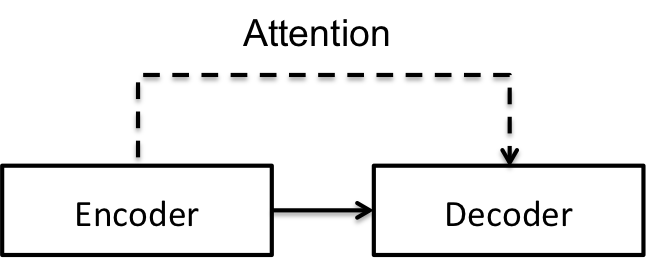}
        \caption{\small Attentive Encoder-Decoder Model.}
        \label{fig:arch-base}
    \end{subfigure}
    ~~~~~~~
    \begin{subfigure}[t]{0.6\textwidth}
        \includegraphics[width=\textwidth]{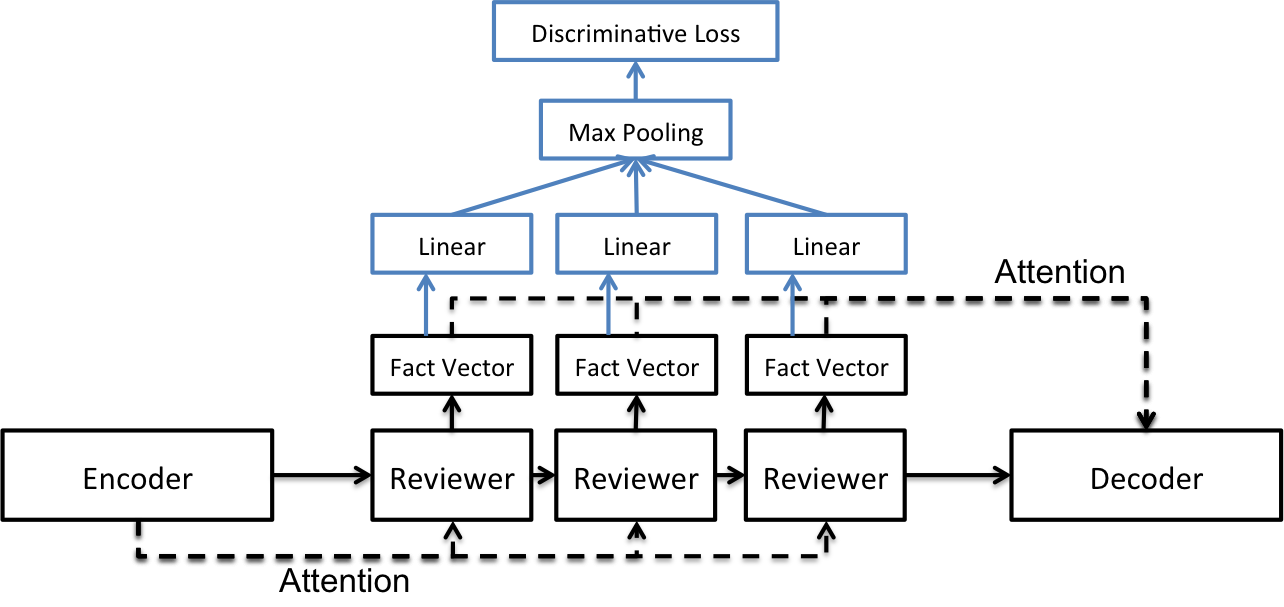}
        \caption{\small Review Network. Blue components denote optional discriminative supervision. $T_r$ is set to $3$ in this example.}
        \label{fig:arch-ours}
    \end{subfigure}
    \caption{\small Model Architectures.}\label{fig:arch}
\end{figure}

In contrast to conventional (attentive) encoder-decoder models, our model consists of three components, encoder, reviewer, and decoder. The comparison of architectures is shown in Figure \ref{fig:arch}. Now we describe the three components in detail.

\subsection{Encoder} \label{sec:cnn}

The encoder encodes the input $\mathbf{x}$ into a context vector $\mathbf{c}$ and a set of hidden states $H = \{\mathbf{h}_t\}_t$. We discuss two types of encoders, RNN encoders and CNN encoders.

{\bf RNN Encoder}: 
Let $T_x = |H|$ be the length of the input sequence. An RNN encoder processes the input sequence $\mathbf{x} = (\mathbf{x}_1, \cdots, \mathbf{x}_{T_x})$ sequentially. At time step $t$, the RNN encoder updates the hidden state by
\[\mathbf{h}_t = f(\mathbf{x}_t, \mathbf{h}_{t - 1}).\]

In this work, we implement $f$ using an LSTM unit. The context vector is defined as the final hidden state $\mathbf{c} = \mathbf{h}_{T_x}$. The cell state and hidden state $\mathbf{h}_0$ of the first LSTM unit are initialized as zero.


{\bf CNN Encoder}: We take a widely-used CNN architecture---VGGNet~\cite{simonyan2014very}---as an example to describe how we use CNNs as encoders. Given a VGGNet, we use the output of the last fully connected layer \texttt{fc7} as the context vector $\mathbf{c} = \mbox{fc7}(\mathbf{x})$, and use $14 \times 14 = 196$ columns of $512$d convolutional output \texttt{conv5} as hidden states $H = \mbox{conv5}(\mathbf{x})$. In this case $T_x = |H| = 196$.

\subsection{Reviewer}

Let $T_r$ be a hyperparameter that specifies the number of review steps. The intuition behind the reviewer module is to review all the information encoded by the encoder and learn thought vectors that are a more compact, abstractive, and global representation than the original encoder hidden states. The reviewer performs $T_r$ review steps on the encoder hidden states $H$ and outputs a thought vector $\mathbf{f}_t$ after each step. More specifically,
\[
\mathbf{f}_t = g_t(H, \mathbf{f}_{t - 1}),
\]
where $g_t$ is a modified LSTM unit with attention mechanism at review step $t$. We study two variants of $g_t$, attentive input reviewers and attentive output reviewers. The attentive input reviewer is inspired by visual attention \cite{xu2015show}, which is more commonly used for images; the attentive output reviewer is inspired by attention on text \cite{bahdanau2014neural}, which is more commonly used for sequential tokens.

\begin{figure}
    \centering
    \begin{subfigure}[t]{0.32\textwidth}
        \includegraphics[width=\textwidth]{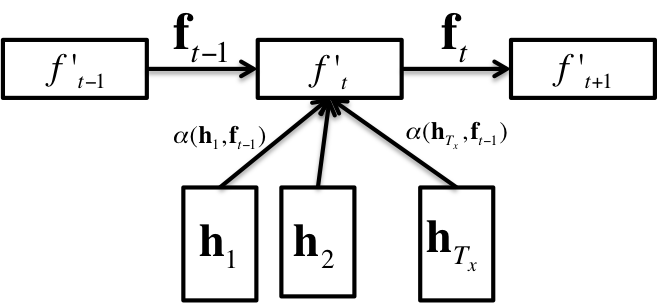}
        \caption{\small Attentive Input Reviewer.}
        \label{fig:review}
    \end{subfigure}
    ~~
    \begin{subfigure}[t]{0.28\textwidth}
        \includegraphics[width=\textwidth]{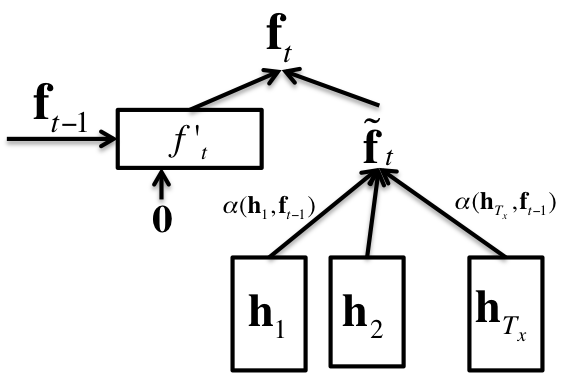}
        \caption{\small Attentive Output Reviewer.}
        \label{fig:review-out}
    \end{subfigure}
    ~~
    \begin{subfigure}[t]{0.32\textwidth}
        \includegraphics[width=\textwidth]{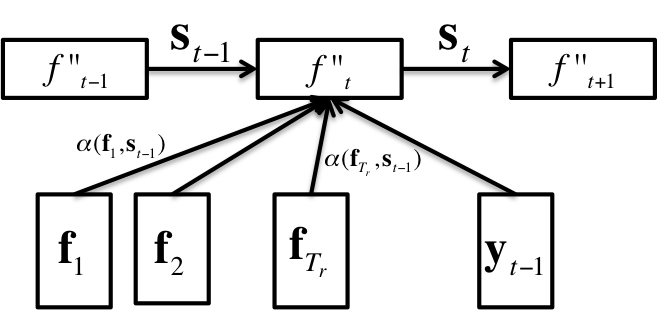}
        \caption{\small Decoder.}
        \label{fig:decode}
    \end{subfigure}
    \caption{\small Illustrations of modules in the review network. $f'_\cdot$ and $f''_\cdot$ denote LSTM units.}\label{fig:module}
\end{figure}


\textbf{Attentive Input Reviewer} At each review step $t$, the attentive input reviewer first applies an attention mechanism on $H$ and use the attention result as the input to an LSTM unit (Cf. Figure \ref{fig:review}). Let $\tilde{\mathbf{f}}_t = \mbox{att}(H, \mathbf{f}_{t - 1})$ be the attention result at step $t$. The attentive input reviewer is formulated as
\begin{equation}
\tilde{\mathbf{f}}_t = \mbox{att}(H, \mathbf{f}_{t - 1}) = \sum_{i = 1}^{|H|} \frac{\alpha(\mathbf{h}_i, \mathbf{f}_{t - 1})}{\sum_{i' = 1}^{|H|} \alpha(\mathbf{h}_{i'}, \mathbf{f}_{t - 1})} \mathbf{h}_i, ~~~ g_t(H, \mathbf{f}_{t - 1}) = f'_t(\tilde{\mathbf{f}}_t, \mathbf{f}_{t - 1}),
\label{eq:att}
\end{equation}
where $\alpha(\mathbf{h}_i, \mathbf{f}_{t - 1})$ is a function that determines the weight for the $i$-th hidden state. $\alpha(\mathbf{x}_1, \mathbf{x}_2)$ can be implemented as a dot product between $\mathbf{x}_1$ and $\mathbf{x}_2$ or a multi-layer perceptron (MLP) that takes the concatenation of $\mathbf{x}_1$ and $\mathbf{x}_2$ as input \cite{luong2015effective}. $f'_t$ is an LSTM unit at step $t$.


\textbf{Attentive Output Reviewer} In contrast to the attentive input reviewer, the attentive output reviewer uses a zero vector as input to the LSTM unit, and the thought vector is computed as the weighted sum of the attention results and the output of the LSTM unit (Cf. Figure \ref{fig:review-out}). More specifically, the attentive output reviewer is formulated as
\[
\tilde{\mathbf{f}}_t = \mbox{att}(H, \mathbf{f}_{t - 1}), ~~~ g_t(H, \mathbf{f}_{t - 1}) = f'_t(\mathbf{0}, \mathbf{f}_{t - 1}) + \mathbf{W} \tilde{\mathbf{f}}_t,
\]
where the attention mechanism \texttt{att} follows the definition in Eq. (\ref{eq:att}), $\mathbf{0}$ denotes a zero vector, $\mathbf{W}$ is a model parameter matrix, and $f'_t$ is an LSTM unit at step $t$. We note that performing attention on top of an RNN unit is commonly used in sequence-to-sequence learning \cite{bahdanau2014neural,luong2015effective,rush2015neural}. We apply a linear transformation with a matrix $\mathbf{W}$ since the dimensions of $f'_t(\cdot, \cdot)$ and $\tilde{\mathbf{f}}_t$ can be different.


\textbf{Weight Tying} We study two variants of weight tying for the reviewer module. Let $\mathbf{w}_t$ denote the parameters for the unit $f'_t$. The first variant follows the common setting in RNNs, where weights are shared among all the units; i.e., $\mathbf{w}_1 = \cdots = \mathbf{w}_{T_r}$. We also observe that the reviewer unit does not have sequential input, so we experiment with the second variant where weights are untied; i.e. $\mathbf{w}_i \not= \mathbf{w}_j, \forall i \not= j$.

The cell state and hidden state of the first unit $f'_1$ are initialized as the context vector $\mathbf{c}$. The cell states and hidden states are passed through all the reviewer units in both cases of weight tying.

\subsection{Decoder}

Let $F = \{\mathbf{f}_t\}_t$ be the set of thought vectors output by the reviewer. The decoder is formulated as an LSTM network with attention on the thought vectors $F$ (Cf. Figure \ref{fig:decode}). Let $\mathbf{s}_t$ be the hidden state of the $t$-th LSTM unit in the decoder. The decoder is formulated as follows:
\begin{equation}
\tilde{\mathbf{s}}_t = \mbox{att}(F, \mathbf{s}_{t - 1}), ~~~ \mathbf{s}_t = f''([\tilde{\mathbf{s}}_t; \mathbf{y}_{t - 1}], \mathbf{s}_{t - 1}), ~~~ y_t = \arg \max_y \mbox{softmax}_y(\mathbf{s}_t),
\label{eq:loss}
\end{equation}
where $[\cdot; \cdot]$ denotes the concatenation of two vectors, $f''$ denotes the decoder LSTM, $\mbox{softmax}_y$ is the probability of word $y$ given by a softmax layer, $y_t$ is the $t$-th decoded token, and $\mathbf{y}_t$ is the word embedding of $y_t$. The attention mechanism \texttt{att} follows the definition in Eq. (\ref{eq:att}). The initial cell state and hidden state $\mathbf{s}_0$ of the decoder LSTM are both set to the \textit{review vector} $\mathbf{r} = \mathbf{W}'[\mathbf{f}_{T_r}; \mathbf{c}]$, where $\mathbf{W}'$ is a model parameter matrix.

\subsection{Discriminative Supervision}

In conventional encoder-decoders, supervision is provided in a generative manner; i.e., the model aims to maximize the conditional probability of generating the sequential output $p(\mathbf{y} | \mathbf{x})$. However, discriminative supervision has been shown to be useful in \cite{fang2015captions}, where the model is guided to predict discriminative objectives, such as the word occurrences in the output $\mathbf{y}$.

We argue that the review network provides a natural way of incorporating discriminative supervision into the model. Here we take word occurrence prediction for example to describe how to incorporate discriminative supervision. As shown in the blue components in Figure \ref{fig:arch-ours}, we first apply a linear layer on top of the thought vector to compute a score for each word at each review step. We then apply a max-pooling layer over all the review units to extract the most salient signal for each word, and add a multi-label margin loss as discriminative supervision. Let $s_{i}$ be the score of word $i$ after the max pooling layer, and $W$ be the set of all words that occur in $\mathbf{y}$. The discriminative loss can be written as
\begin{equation}
\mathcal{L}_d = \frac{1}{Z} \sum_{j \in W} \sum_{i \not= j} \max(0, 1 - (s_j - s_i)),
\label{eq:d-loss}
\end{equation}
where $Z$ is a normalizer that counts all the valid $i, j$ pairs. We note that when the discriminative supervision is derived from the given data (i.e., predicting word occurrences in captions), we are not using extra information.

\subsection{Training}

The training loss for a single training instance $(\mathbf{x}, \mathbf{y})$ is defined as a weighted sum of the negative conditional log likelihood and the discriminative loss. Let $T_y$ be the length of the output sequence $\mathbf{y}$. The loss can be written as
\[
\mathcal{L}(\mathbf{x}, \mathbf{y}) = \frac{1}{T_y} \sum_{t = 1}^{T_y} - \log \mbox{softmax}_{y_t}(\mathbf{s}_t) + \lambda \mathcal{L}_d,
\]
where the definition of $\mbox{softmax}_y$ and $\mathbf{s}_t$ follows Eq. (\ref{eq:loss}), and the formulation of $\mathcal{L}_d$ follows Eq. (\ref{eq:d-loss}). $\lambda$ is a constant weighting factor. We adopt adaptive stochastic gradient descent (AdaGrad) \cite{duchi2011adaptive} to train the model in an end-to-end manner. The loss of a training batch is averaged over all instances in the batch.

\subsection{Connection to Encoder-Decoders}

We now show that our model can be reduced to the conventional (attentive) encoder-decoders in a special case. In attentive encoder-decoders, the decoder takes the context vector $\mathbf{c}$ and the set of encoder hidden states $H = \{\mathbf{h}_t\}_t$ as input, while in our review network, the input of the decoder is instead the review vector $\mathbf{r}$ and the set of thought vectors $F = \{\mathbf{f}_t\}_t$. To show that our model can be reduced to attentive encoder-decoders, we only need to construct a case where $H = F$ and $\mathbf{c} = \mathbf{r}$.

Since $\mathbf{r} = \mathbf{W}'[\mathbf{f}_{T_r}; \mathbf{c}]$, it can be reduced to $\mathbf{r} = \mathbf{c}$ with a specific setting of $\mathbf{W}'$. We further set $T_r = T_x$, and define each reviewer unit as an identity mapping $g_t(H, \mathbf{f}_{t - 1}) = \mathbf{h}_t$, which satisfies the definition of both the attentive input reviewer and the attentive output reviewer with untied weights. With the above setting, we have $\mathbf{h}_t = \mathbf{f}_t, \forall t = 1, \cdots, T_x$; i.e., $H = F$. Thus our model can be reduced to attentive encoder-decoders in a special case. Similarly we can show that our model can be reduced to vanilla encoder-decoders (without attention) by constructing a case where $\mathbf{r} = \mathbf{c}$ and $\mathbf{f}_t = \mathbf{0}$. Therefore, our model is more expressive than (attentive) encoder-decoders.

Though we set $T_r = T_x$ in the above construction, in practice, we set the number of review steps $T_r$ to be much smaller compared to $T_x$, since we find that the review network can learn a more compact and effective representation.

\section{Experiments}

We experiment with two datasets of different tasks, image captioning and source code captioning. Since these two tasks are quite different, we can use them to test the robustness and generalizability of our model.

\subsection{Image Captioning}

\subsubsection{Offline Evaluation}

We evaluate our model on the MSCOCO benchmark dataset \cite{chen2015microsoft} for image captioning. The dataset contains 123,000 images with at least 5 captions for each image. For offline evaluation, we use the same data split as in \cite{karpathy2015deep,xu2015show,you2016image}, where we reserve 5,000 images for development and test respectively and use the rest for training. The models are evaluated using the official MSCOCO evaluation scripts. We report three widely used automatic evaluation metrics, BLEU-4, METEOR, and CIDEr.

We remove all the non-alphabetic characters in the captions, transform all letters to lowercase, and tokenize the captions using white space. We replace all words occurring less than 5 times with an unknown token \texttt{<UNK>} and obtain a vocabulary of 9,520 words. We truncate all the captions longer than 30 tokens.

We set the number of review steps $T_r = 8$, the weighting factor $\lambda = 10.0$, the dimension of word embeddings to be $100$, the learning rate to be $1\mathrm{e}{-2}$, and the dimension of LSTM hidden states to be $1,024$. These hyperparameters are tuned on the development set. We also use early stopping strategies to prevent overfitting. More specifically, we stop the training procedure when the BLEU-4 score on the development set reaches the maximum. We use an MLP with one hidden layer of size $512$ to define the function $\alpha(\cdot, \cdot)$ in the attention mechanism, and use an attentive input reviewer in our experiments to be consistent with visual attention models~\cite{xu2015show}. We use beam search with beam size 3 for decoding. We guide the model to predict the words occurring in the caption through the discriminative supervision $\mathcal{L}_d$ without introducing extra information. We fix the parameters of the CNN encoders during training.



\begin{table}[t]
  \caption{\small Comparison of model variants on MSCOCO dataset. Results are obtained with a single model using VGGNet. Scores in the brackets are without beam search. We use RNN-like tied weights for the review network unless otherwise indicated. ``Disc Sup'' means discriminative supervision.}
  \label{tab:var}
  \centering
  \begin{tabular}{llll}
    \toprule
    Model & BLEU-4 & METEOR & CIDEr \\
    \midrule
    Attentive Encoder-Decoder & 0.278 (0.255) & 0.229 (0.223) & 0.840 (0.793) \\
    \midrule
    Review Net & 0.282 (0.259) & 0.233 (0.227) & 0.852 (0.816) \\
    Review Net + Disc Sup & 0.287 (0.264) & \textbf{0.238} (\textbf{0.232}) & 0.879 (0.833) \\
    Review Net + Disc Sup + Untied Weights & \textbf{0.290} (\textbf{0.268}) & 0.237 (\textbf{0.232}) & \textbf{0.886} (\textbf{0.852}) \\
    \bottomrule
  \end{tabular}
\vspace{-0.1in}
\end{table}

We compare our model with encoder-decoders to study the effectiveness of the review network. We also compare different variants of our model to evaluate the effects of different weight tying strategies and discriminative supervision. Results are reported in Table \ref{tab:var}. All the results in Table \ref{tab:var} are obtained using VGGNet \cite{simonyan2014very} as encoders as described in Section \ref{sec:cnn}.

From Table \ref{tab:var}, we can see that the review network can improve the performance over conventional attentive encoder-decoders consistently on all the three metrics.
We also observe that adding discriminative supervision can boost the performance, which demonstrates the effectiveness of incorporating discriminative supervision in an end-to-end manner.
Untying the weights between the reviewer units can further improve the performance. Our conjecture is that the models with untied weights are more expressive than shared-weight models since each unit can have its own parametric function to compute the thought vector. In addition to Table \ref{tab:var}, our experiment shows that applying discriminative supervision on attentive encoder-decoders can improve the CIDEr score from $0.793$ to $0.811$ without beam search. We did experiments on the development set with $T_r$ = 0, 4, 8, and 16. The performances when $T_r = 4$ and $T_r = 16$ are slightly worse then $T_r = 8$ ($-0.003$ in Bleu-4 and $-0.01$ in CIDEr). We also experimented on the development set with $\lambda$ = 0, 5, 10, and 20, and $\lambda = 10$ gives the best performance.

\subsubsection{Online Evaluation on MSCOCO Server}

\begin{table}[t]
  \caption{\small Comparison with state-of-the-art systems on the MSCOCO evaluation server. $\dagger$ indicates ensemble models. \textit{Feat.} means using task-specific features or attributes. \textit{Fine.} means using CNN fine-tuning.}
  \label{tab:sota}
  \centering
  \begin{tabular}{lllllll}
    \toprule
    Model & BLEU-4 & METEOR & ROUGE-L & CIDEr & Fine. & Feat. \\
    \midrule
    Attention \cite{xu2015show} & 0.537 & 0.322 & 0.654 & 0.893 & No & No \\
    MS Research \cite{fang2015captions} & 0.567  & 0.331 & 0.662 & 0.925 & No & Yes \\
    Google NIC \cite{vinyals2015show}$^{\dagger}$ & 0.587 & 0.346 & 0.682 & 0.946 & Yes & No  \\
    Semantic Attention \cite{you2016image}$^{\dagger}$ & \textbf{0.599} & 0.335  & 0.682 & 0.958 & No & Yes \\
    \midrule
    Review Net (this paper)$^{\dagger}$ & 0.597 & \textbf{0.347} & \textbf{0.686} & \textbf{0.969} & No & No \\
    \bottomrule
  \end{tabular}
\end{table}

We also compare our model with state-of-the-art systems on the MSCOCO evaluation server in Table \ref{tab:sota}. Our submission uses Inception-v3 \cite{szegedy2015rethinking} as the encoder and is an ensemble of three identical models with different random initialization. We take the output of the last convolutional layer (before pooling) as the encoder states. From Table \ref{tab:sota}, we can see that among state-of-the-art published systems, the review network achieves the best performance for three out of four metrics (i.e., METEOR, ROUGE-L, and CIDEr), and has very close performance to Semantic Attention \cite{you2016image} on BLEU-4 score.

The Google NIC system \cite{vinyals2015show} employs several tricks such as CNN fine-tuning and scheduled sampling and takes more than two weeks to train; the semantic attention system requires hand-engineering task-specific features/attributes. Unlike these methods, our approach with the review network is a generic end-to-end encoder-decoder model and can be trained within six hours on a Titan X GPU.


\subsubsection{Case Study and Visualization}

\begin{figure}
    \centering
    \includegraphics[width=\textwidth]{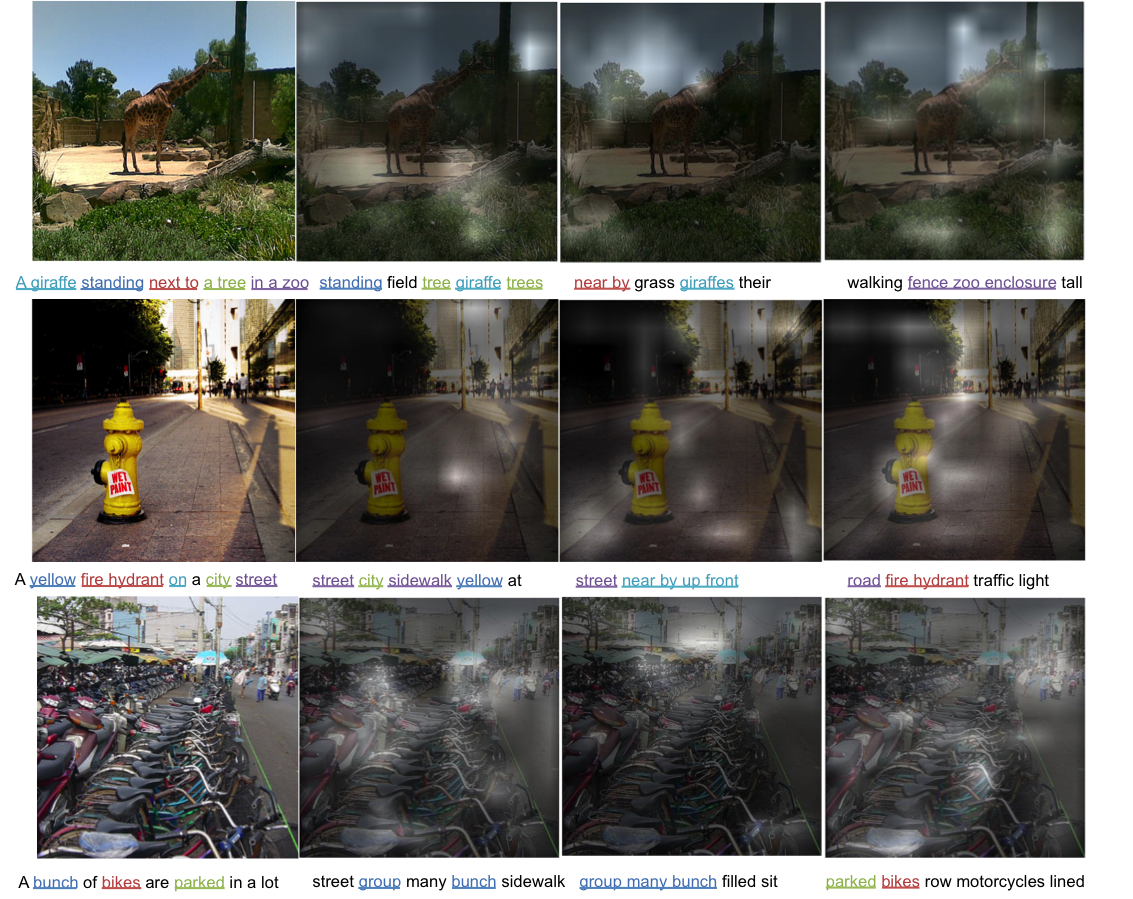}
\vspace{-0.1in}
    \caption{\small Each row corresponds to a test image: the first is the original image with the caption output by our model, and the following three images are the visualized attention weights of the first three reviewer units. We also list the top-5 words with highest scores for each unit. Colors indicate semantically similar words.}\label{fig:vis}
\vspace{-0.1in}
\end{figure}

To better understand the review network, we visualize the attention weights $\alpha$ in the review network in Figure \ref{fig:vis}. The visualization is based on the review network with untied weights and discriminative supervision. We also list the top-5 words with highest scores (computed based on the thought vectors) at each reviewer unit.

We find that the top words with highest scores can uncover the reasoning procedure underlying the review network. For example, in the first image (a giraffe in a zoo), the first reviewer focuses on the motion of the giraffe and the tree near it, the second reviewer analyzes the relative position between the giraffe and the tree, and the third reviewer looks at the big picture and infers that the scene is in a zoo based on recognizing the fences and enclosures. All the above information is stored in the thought vectors and decoded as natural language by the decoder.

Different from attentive encoder-decoders \cite{xu2015show} that attend to a single object at a time during generation, it can be clearly seen from Figure \ref{fig:vis} that the review network captures more global signals, usually combining multiple objects into one thought, including objects not finally shown in the caption (e.g., ``traffic light'' and ``motorcycles''). The thoughts are sometimes abstractive, such as motion (``standing''), relative position (``near'', ``by'', ``up''), quantity (``bunch'', ``group''), and scene (``city'', ``zoo''). Also, the order of review is not restricted by the order in natural language.



\subsection{Source Code Captioning}

\subsubsection{Data and Settings}

The task of source code captioning is to predict the code comment given the source code, which can be framed under the problem of sequence-to-sequence learning. We experiment with a benchmark dataset for source code captioning, HabeasCorpus \cite{movshovitz2013natural}. HabeasCorpus collects nine popular open-source Java code repositories, such as Apache Ant and Lucene. The dataset contains $6,734$ Java source code files with $7,903,872$ source code tokens and $251,565$ comment word tokens. We randomly sample 10\% of the files as the test set, 10\% as the development set, and use the rest for training. We use the development set for early stopping and hyperparameter tuning.

Our evaluation follows previous works on source code language modeling \cite{maddison2014structured} and captioning \cite{movshovitz2013natural}. We report the log-likelihood of generating the actual code captions based on the learned models. We also evaluate the approaches from the perspective of code comment completion, where we compute the percentage of characters that can be saved by applying the models to predict the next token. More specifically, we use a metric of \textit{top-$k$ character savings} \cite{movshovitz2013natural} (CS-$k$). Let $n$ be the minimum number of prefix characters needed to be filtered such that the actual word ranks among the top-$k$ based on the given model. Let $L$ be the length of the actual word. The number of saved characters is then $L - n$. We compute the average percentage of saved characters per comment to obtain the metric CS-$k$.

We follow the tokenization used in \cite{movshovitz2013natural}, where we transform camel case identifiers into multiple separate words (e.g., ``binaryClassifierEnsemble'' to ``binary classifier ensemble''), and remove all non-alphabetic characters. We truncate code sequences and comment sequences longer than 300 tokens.
We use an RNN encoder and an attentive output reviewer with tied weights. We set the number of review steps $T_r = 8$, the dimension of word embeddings to be 50, and the dimension of the LSTM hidden states to be 256.

\subsubsection{Results}

\begin{table}[t]
  \caption{\small Comparison of model variants on HabeasCorpus code captioning dataset. ``Bidir'' indicates using bidirectional RNN encoders, ``LLH'' refers to log-likelihood, ``CS-$k$'' refers to top-$k$ character savings.}
  \label{tab:code}
  \centering
  \begin{tabular}{lllllll}
    \toprule
    Model & LLH & CS-$1$ & CS-$2$ & CS-$3$ & CS-$4$ & CS-$5$ \\
    \midrule
    Language Model & -5.34 & 0.2340 & 0.2763 & 0.3000 & 0.3153 & 0.3290 \\
    Encoder-Decoder & -5.25 & 0.2535 & 0.2976 & 0.3201 & 0.3367 & 0.3507 \\
    Encoder-Decoder (Bidir) & -5.19 & 0.2632 & 0.3068 & 0.3290 & 0.3442 & 0.3570 \\
    Attentive Encoder-Decoder (Bidir) & -5.14 & 0.2716 & 0.3152 & 0.3364 & 0.3523 & 0.3651 \\
    \midrule
    Review Net & \textbf{-5.06} & \textbf{0.2889} & \textbf{0.3361} & \textbf{0.3579} & \textbf{0.3731} & \textbf{0.3840} \\
    \bottomrule
  \end{tabular}
\vspace{-0.1in}
\end{table}

We report the log-likelihood and top-$k$ character savings of different model variants in Table \ref{tab:code}. The baseline model ``Language Model'' is an LSTM decoder whose output is not sensitive to the input code sequence. A preliminary experiment showed that the LSTM decoder significantly outperforms the N-gram models used in \cite{movshovitz2013natural} (+3\% in CS-$2$), so we use the LSTM decoder as a baseline for comparison. We also compare with different variants of encoder-decoders, including incorporating bidirectional RNN encoders and attention mechanism. It can be seen from Table \ref{tab:code} that both bidirectional encoders and attention mechanism can improve over vanilla encoder-decoders. The review network outperforms attentive encoder-decoders consistently in all the metrics, which indicates that the review network is effective at learning useful representation.

\section{Conclusion}

We present a novel architecture, the review network, to improve the encoder-decoder learning framework. The review network performs multiple review steps with attention on the encoder hidden states, and computes a set of thought vectors that summarize the global information in the input. We empirically show consistent improvement over conventional encoder-decoders on the tasks of image captioning and source code captioning. In the future, it will be interesting to apply our model to more tasks that can be modeled under the encoder-decoder framework, such as machine translation and text summarization.


\textbf{Acknowledgements} This work was funded by the NSF under grants CCF-1414030 and IIS-1250956, Google, Disney Research, the ONR grant N000141512791, and the ADeLAIDE grant FA8750-16C-0130-001.

\bibliographystyle{plain}
\bibliography{nips_2016}

\end{document}